# Using Explainable AI to Cross-Validate Socio-economic Disparities Among Covid-19 Patient Mortality

Li Shi[1], Redoan Rahman[1], Esther Melamed, MD, PhD[2], Jacek Gwizdka, PhD[1], Justin F. Rousseau, MD, MMSc[2*], Ying Ding, PhD[1*]

[1]School of Information, University of Texas at Austin, Austin, Texas, USA; [2]Dell Medical School, University of Texas at Austin, Austin, Texas, USA

*joint senior authors

**Abstract**

*This paper applies eXplainable Artificial Intelligence (XAI) methods to investigate the socioeconomic disparities in COVID-19 patient mortality. An Extreme Gradient Boosting (XGBoost) prediction model is built based on a de-identified Austin area hospital dataset to predict the mortality of COVID-19 patients. We apply two XAI methods, Shapley Additive exPlanations (SHAP) and Locally Interpretable Model Agnostic Explanations (LIME), to compare the global and local interpretation of feature importance. This paper demonstrates the advantages of using XAI which shows the feature importance and decisive capability. Furthermore, we use the XAI methods to cross-validate their interpretations for individual patients. The XAI models reveal that Medicare financial class, older age, and gender have high impact on the mortality prediction. We find that LIME's local interpretation does not show significant differences in feature importance comparing to SHAP, which suggests pattern confirmation. This paper demonstrates the importance of XAI methods in cross-validation of feature attributions.*

## 1 Introduction

Socioeconomic status, such as health care access, race, and ethnicity, profoundly impact COVID-19 incidence and mortality [1]. Recent meta-review of 4.3 million patients from 68 studies highlights the strong association between socioeconomic disparity and COVID-19 mortality [2]. Understanding the effect of such societal factors helps us develop better health policies and health delivery methods to mitigate the disparity. Different works of literature show a strong relationship between socio-economic status and mortality and highlight the importance of socio-economic inequalities in health outcomes [3]. However, previous studies on socioeconomic features could not provide a comprehensive explanation from the patient cohort level and at the individual patient level (3). Especially, cross-validation of different explanation methods is critical because explanation offered by one method can be biased and one-sided.

EXplainable Artificial Intelligence (XAI) methods provide a new way to understand feature importance. They offer a variety of perspectives to dissect feature importance to highlight the decisive capability of each feature, the dependence of different features, and the joint contributions to illustrate how these features shift the prediction of an individual patient either towards the mean or away from the mean [4]. Therefore, XAI methods offer the unique advantages than conventional feature importance in machine learning methods. Here, we apply two XAI methods, Shapley Additive exPlanations (SHAP) and Locally Interpretable Model Agnostic Explanations (LIME), to investigate the effect of socioeconomic disparities on COVID-19 patient mortality. Unlike previous XAI approaches [5,6], our methods provide several improvements including decisive capability identification, individual local interpretation, and cross-validation between different XAI methods.

This paper examines a hospital dataset from Austin containing information regarding COVID-19 positive patients from March 2020 to December 2021. We apply the Extreme Gradient Boosting (XGBoost) algorithm to the dataset to understand the impact of the socioeconomic features on the patient mortality prediction. We use the Shapley score to illustrate the decisive capability of each feature. Furthermore, we use SHAP and LIME to analyze individual patient mortality prediction, which allows us to understand how individual feature contributions differ from the global contributions. Finally, we propose statistical methods to cross-validate the local interpretations from SHAP and LIME. Our study emphasizes the importance of socioeconomic features in COVID-19 mortality, and identifies that Medicare financial class, older age, and gender have high impact on the mortality prediction for COVID-19 patients in Austin.

## 2 Background

XAI enables humans to interpret the model prediction process and enhance confidence in the result applications. Therefore, XAI tools have become more prevalent in healthcare analysis, especially in COVID-19 related studies over the past few years. **[7]** used Shapley global interpretation to indicate socio-economic factors' importance in confirmed case rate and death rate. In this research, non-white poverty has a high positive impact on the death rate while uninsured plays a relatively high negative impact on the death rate in the southern states. **[8]** proposed a web-based architecture based on Random Forest and XGBoost classifiers and then used Shapley values to demonstrate the varying dependence of various health factors on predicting the COVID-19 risk level. **[9]** analyzed the relationship between clinical data and COVID-19 mortality. In this research, the authors applied LIME to illustrate that HbA1c (i.e., a marker of diabetes control) was the most significant contributor to the mortality risk. Among all these studies, the researchers used feature importance plots to demonstrate various factors' contribution to the model prediction. However, in these pieces of literature, the interpretations did not consider the different impacts of the factors when they have different values.

Other studies used the SHAP summary plot to understand the decisive factor capacities. **[10]** predicted the admission to the ICU and mortality across the COVID-19 patients who received heparin by using the Hybrid Extreme Gradient Boosting (HXGBoost) classifier. The research used a SHAP summary plot to interpret the HXGBoost model and found that low lymphocyte count at day seven combined with increased FiO2 on days 1 and 5 increase the risk of mortality. **[11]** assessed the contribution of the impulse-radio ultra-wideband radar factors to the predictions of COVID-19 test results with the help of SHAP summary and feature importance plots and then used the analysis results for feature selection. In these research studies, the involved researchers used SHAP summary plots to understand the decisive power of the features. This XAI method could help people to understand how the variation of the factor values would impact the prediction outcome. This paper will use SHAP summary plot to investigate how the existence or the absence of particular socio-economic categories would impact the prediction of patient mortality.

XAI was not only used to understand how factors impact the predictions globally but also used to understand the roles that the factors played in each instance. **[12]** used a random forest model to dynamically predict the subsequent day mortality risk of COVID-19 patients based on summary characteristics of longitudinal risk factor trajectories. They used SHAP local interpretation to compare the dynamic factor contributions and prediction outcomes on COVID-19-positive survivors versus COVID-19 positive non-survivor. **[13]** used the Shapley force plot to interpret how five core laboratory parameters influence true positive and false negative predictions locally. **[4]** interpreted the random forest classifier using the Shapley waterfall plot, showing the local feature importance for the individuals who died due to COVID-19 with or without diabetes mellitus. The evidence that individual feature contributions differed from global contributions stressed the advantages of individualized risk explanations over generic risk descriptions. All the previous research findings using local interpretation illustrated that high local impact features may not have a high impact globally. Therefore, local interpretation could provide a unique view of how the factors work in different circumstances. In this research, we will use local interpretation to learn how socio-economic factors work in individual examples and therefore shed light on the practical applications of individualized mortality risk explanations.

Previous research demonstrated that there is more than one tool in the XAI method and each tool has its own computational mechanisms and interpretation method. Therefore, cross-validation could increase the reliability of the model interpretations. Researchers mostly use SHAP for global interpretation while LIME for local interpretation **[6,14]**. In **[5]**, the authors used SHAP and LIME for global and local interpretation to understand the impact of socio-economic factors on COVID-19 mortality. They visualized the global and local interpretations, then briefly compared the ranks of the factors for cross-validation and concluded that age is the most important feature and time to hospital after symptom onset as the second most important feature to the mortality predictions. Even though this research compares the results from SHAP and LIME correspondingly, there was not a cross-validation method in this research using statistical methods. Therefore, our research would use SHAP and LIME to interpret the model predictions and utilize statistical methods to provide solid cross-validations to increase the reliability and trustworthiness of the machine learning model interpretations.

## 3 Methods
### 3.1 LIME (Locally Interpretable Model Agnostic Explanations)
The LIME method provides interpretations for the individual models by locally approximating the model around the given prediction. It is a post-hoc model-agnostic explanation technique. LIME is independent of the original classifier and works on specific observations [15]. LIME tries to fit a local model using sample data points similar to the explained observation data points. LIME produces the explanations as follows:

$$\xi(x) = \underset{g \in G}{\operatorname{argmin}} \mathcal{L}(f, g, \pi_x) + \Omega(g)$$

where G is a class of potentially interpretable models, $g \in G$ is an explanation as a model, $\Omega(g)$ is a measure of complexity, $\pi_x$ is a proximity measure between an instance z to x, and f denotes $\mathbb{R}^d \rightarrow \mathbb{R}$. LIME aims to minimize the locality aware loss $\mathcal{L}$ without making assumptions about f.

### 3.2 SHAP (Shapley Additive exPlanations)

SHAP is an integrated framework that explains the output of any model based on Shapley values. It measures the feature importance for linear models in the presence of multicollinearity[16]. In contrast to LIME, SHAP does not have to build a local model. The Shapley values are calculated from coalitional game theory and express model predictions as combinations of binary variables representing the existence of each covariate. The model is retrained on all feature subsets $S \subseteq F$, then predictions are weighed against the presence and missingness of the variable. Thus, Shapley values are calculated from the computed values and used as feature importance which is a weighted average of all possible references:

$$\Phi_i = \sum_{S \subseteq F\{i\}} \frac{|S|!\,(|F| - |S| - 1)!}{|F|} f_{S \cup \{i\}}(x_{S \cup \{i\}}) - f_s(x_s)$$

Here, $\Phi_i$ represents the feature importance of the i-th feature. This value can be positive or negative depending on the impact of the feature on the model prediction. If the feature impacts the model positively, then the assigned Shapley value to the feature is positive, and if the effect is negative, then the Shapley value reflects that impact.

### 3.3 Cross Validation between SHAP and LIME

In general, LIME calculates the difference between the predictions with or without the variable, while SHAP measures the variable's contribution to the difference between the actual and mean predictions. Due to differences in the computational mechanisms, the local interpretation of SHAP and LIME may reveal the features to have different impact rankings and tendencies contributing to different categories. To compare the local interpretation overall performance of SHAP and LIME, we measure the feature impact consistency and ranking difference to quantify the difference between SHAP and LIME interpretation.

*Impact Value Consistency*. When SHAP and LIME values are positive, the features push the prediction to mortality. When negative, the features push the prediction to non-mortality. Therefore, we compare the consistency of the contribution tendency by comparing the signs of the SHAP and LIME values. Then we calculate the ratio of the consistent features' impact on each instance's prediction tendency.

$$ratio\ of\ impact\ value\ consistency = \frac{\#\ of\ features\ with\ same\ signs\ in\ SHAP\ and\ LIME}{\#\ of\ the\ feature\ used\ in\ model\ prediction}$$

*Impact Ranking Difference*. We selected features with consistent contribution tendencies and compared their ranking differences between SHAP and LIME. For each instance, the features are separated based on the sign of the feature impact values and then given a rank based on the absolute feature impact value (i.e., the feature with the highest impact assigned as 1). Then we filtered out the features with consistent contribution tendency and calculated the rank difference for each feature.

$$impact\ ranking\ difference = SHAP\ feature\ ranking - LIME\ feature\ ranking$$

### 3.4 Dataset and Experiment Setup

Electronic health record (EHR) data for 12,740 patients diagnosed with COVID-19 from March 2020 to December 2021 were extracted from the Ascension Seton Hospital Network clinical data warehouse with support and data sharing processes of the Dell Medical School Enterprise Data Intelligence Data Core [17]. Data were collected for five distinct hospitals comprised of three community hospitals, an academic hospital, and a children's hospital in Central Texas. Each patient visit was recorded as a data point in the dataset and in total there are 20,180 COVID-19-positive patient visits. The data were de-identified with all clinical events indexed to the first admission with a confirmed COVID-19 diagnosis (U07.1). This study was conducted according to the guidelines of the Declaration of Helsinki and approved by the Institutional Review Board of The University of Texas at Austin (IRB ID: 2020-04-0117).

For the experiment, we selected ten different features to predict the mortality of COVID-19 patients. Table 1 describes the features in detail. To better understand how the features impact the predictions, we encoded most of the categorical data with OneHotEncoder by encoding the categorical features as one-hot numeric arrays and split each category as individual columns. However, considering the admission source feature has too many categories, we remained using the OrdinalEncoder scaling technique to encode it. For the model training, we excluded the data if it had missing values in any column. In the filtered dataset, 18,368 patients tested positive, among which 601 patients died. The data was split into 80% and 20% for training and testing, respectively.

In this research, we chose Extreme Gradient Boosting (XGBoost) as our machine learning model. It is a highly scalable tree boosting system that is sparsity-aware and applicable in different scenarios. XGBClassifier was used from the xgboost python library and set the objective as 'reg:logistic' since we are working on a binary classification problem. We tuned the parameters to prevent overfitting and finally set the 'learning_rate' as 0.1, 'max_depth' as 5, 'n_estimators' as 10. Considering the data were imbalanced between two classes, 'scale_pos_weight' was set as 'sqrt(total number of negative examples/total number of positive examples)' to assign different weights to the non-mortality and mortality classes. After training the model, we used the Shapley feature importance and summary plot to analyze the decisive power of the factors. Then we selected one mortality class individual and one non-mortality class individual to understand the local feature importance similarity and difference between SHAP and LIME. Then, we cross-validate the interpretations from SHAP and LIME by comparing the impact value consistency and impact ranking differences of SHAP and LIME. All the coding was done in the Jupyter Notebook.

**Table 1.** Selected feature explanations and Encoding methods.

| Feature name | Meaning | Encoding method |
| --- | --- | --- |
| Encounter type | Classes to distinguish between different health care settings | *OneHotEncoder* (encnt_Emergency, encnt_Outpatient, encnt_Inpatient) |
| Admission source | The origin of the patient's admission to the hospital | *Ordinal Encoder* (0 = clinic referral, 1 = court/law enforcement, 2 = emergency room, 3 = HMO (health maintenance organization) referral, 4 = newborn (extramural birth), 5 = newborn (normal delivery), 6 = physician referral, 7 = transfer from a hospital, 8 = transfer from a skilled nursing facility, 9 = transfer from another health care facility (includes rehabilitation and psychiatric facilities)) |
| Race | Individual's race | *OneHotEncoder* (race_BlackAfricanAmerican, race_White, race_Other) |
| Ethnicity | Individual's cultural identification | *OneHotEncoder* (ethnicity_HispanicLatino, ethnicity_Not) |
| Gender | Individual's gender | *OneHotEncoder* (gender_F, gender_M) |
| Financial class | Payor groups for billing and reporting purposes | *OneHotEncoder* (financ_Medicare, financ_Medicaid, financ_Self, financ_Commercial) |
| Age | Individual's age | |
| Deidentified zip code | First 3 digits of the patient resident zip code | |
| Admit quarter | Quarter when the individual was admitted | |
| Admit year | Year when the individual was admitted | |

## 4 Results

XGBoost model achieved a good performance (precision (positive predictive value) = 0.95, recall (sensitivity)=0.97) in predicting mortality of COVID-19 patients based on their socio-economic information (Table 2). We employed SHAP and LIME tools to understand how the features impact the prediction globally and locally.

**Table 2.** Classification report for XGBoost model predictions on test set.

|                  | Precision | Recall | F1 Score | Support |
|------------------|-----------|--------|----------|---------|
| 0 (non-mortality) | 0.97      | 1.00   | 0.98     | 3556    |
| 1 (mortality)    | 0.50      | 0.03   | 0.05     | 118     |
| Accuracy         |           |        | 0.97     | 3674    |
| Macro Avg        | 0.73      | 0.51   | 0.52     | 3674    |
| Weighted Avg     | 0.95      | 0.97   | 0.95     | 3674    |

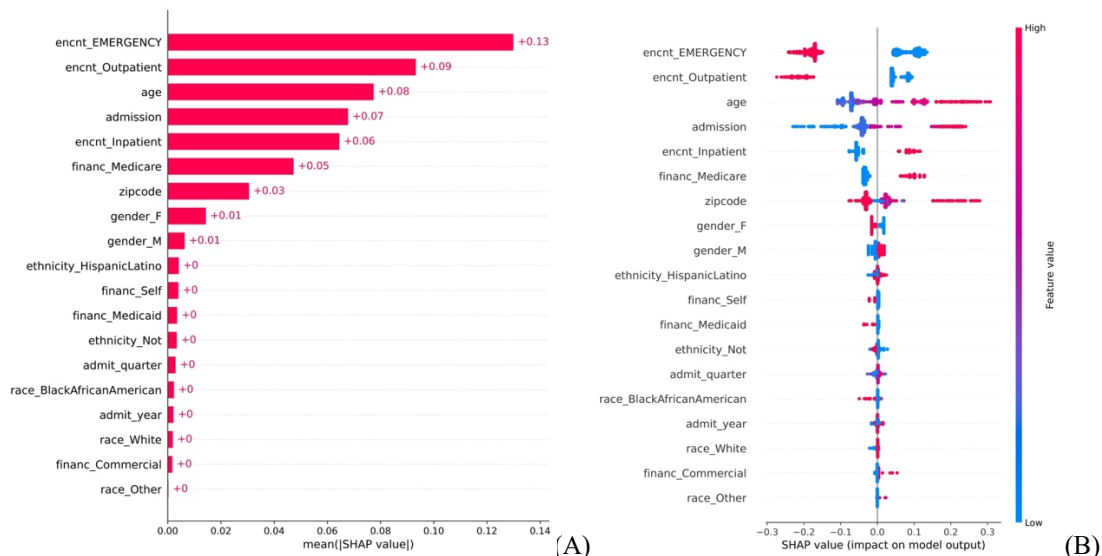

**Figure 1.** (A) SHAP feature importance plot and (B) SHAP summary plot.

### 4.1 Feature decisive power

We utilize SHAP feature importance plot and summary plot to understand how the impact of socio-economic features on mortality predictions. The SHAP feature importance plot (Figure 1 (a)) illustrates the average impact (i.e., the average absolute value of SHAP) of each feature. The higher the ranking, the more significant the feature's impact on the mortality prediction model. It indicates that patient encounter type (i.e., emergency, outpatient, inpatient) plays a critical role in identifying patient mortality. Age, admission type, Medicare financial class, and gender also play a decisive role in the prediction. The other financial classes, ethnicity, race, and admitted quarter and year have little effect on the output.

The SHAP summary plot (Figure 1(b)) reveals how the decisive capacities of each feature change according to each value. Features are sorted according to global contribution. The color of the dots in each feature row represents the instance feature value and the horizontal positions represents the instance SHAP value. The clustering of the same color dots indicates the possible relationship between feature value, and the impact on the model prediction. The figure shows that when the encounter type is emergency or outpatient (i.e., feature value=1), it has a decisive power towards non-mortality prediction. In contrast, if the encounter type is not emergency or outpatient (i.e., feature value =0), it has a relatively low power towards mortality prediction. Correspondingly, patients tend to have a higher probability of dying when the encounter type is inpatient. Different ages also have different power in the prediction. When age is higher, they are more likely to suffer mortality. However, when age is lower, the mixed pattern shows that younger age mostly contributes to the non-mortality but they might have either a high or a low impact on the predictions. The pattern in the admission source feature indicates that when the patients are more likely to die when are transferred from other medical facilities. Medicare financial class ranks sixth and illustrates that patients with Medicare insurance are more likely to die. However, when the financial class is Medicaid, patients are slightly less likely to die. The figure also illustrates that the female gender contributes to the non-mortality output, while the male gender contributes to the mortality output. Other features' SHAP values are primarily gathered around 0, showing that they do not contribute substantially to the mortality prediction.

### 4.2 Local interpretation

Local interpretations allow us to understand how the features affect the individual prediction and how the feature contributions are consistent with or different from the global average in specific circumstances. Figures 2 and 3 show the local interpretation of SHAP and LIME from one non-mortal patient prediction (true negative). Figure 2(A) and 2(B) are SHAP local interpretation plots. Features in red are pushing the prediction toward mortality, while the ones in blue push non-mortality. The size of the bar indicates impact power on the predictions. This figure reveals that, for this individual case, age at 75, financial class as Medicare, zip code starting with 786, and non-female gender contribute to the mortality prediction. On the other hand, encounter type as emergency, admission type as emergency room, and admit year as 2021 pushes contribute to the higher survival rate. Figure 3 shows the LIME local interpretation summarizing the factors contributing to the prediction, where orange-colored factors indicate the contribution to mortality prediction and blue-colored factors indicate the contribution to non-mortality prediction. The table on the right of Figure 3 shows the contribution's rank with the feature's actual value for the specific instance. It shows that emergency encounter type and admission source have the most significant impact on the prediction of non-mortality, followed by not-outpatient encounter type, age, and Medicare financial type contributing to the mortality prediction. Same as SHAP local interpretation, race, admit year, and admit quarter have a low impact on the prediction. For this specific individual, we found that Medicare financial class plays an essential role in the mortality-class prediction, which can hardly be seen from the global interpretation since the older age is more polarized than Medicare in the SHAP summary plot.

Figure 4 and 5 show local interpretation of SHAP and LIME from one mortal patient prediction (true positive). Figure 4 illustrates that, for this individual case, admission source as 'transfer from another health care facility', zip code being 789, encounter type as inpatient, financial class as Medicare, age at 64 contribute to the mortality prediction. LIME interpretation in Figure 5 shows that zip code, encounter type, admission source, age, financial class have larger impact on the prediction towards mortality. In both figures, zip code, encounter type, admission source, financial class, and age rank high and push the prediction to higher probability of mortality in the local interpretation. Even though the SHAP summary plot indicates that zip code as a medium-high impactful feature, in this individual case, zip code is in top 2 ranked feature contributing to the mortality prediction.

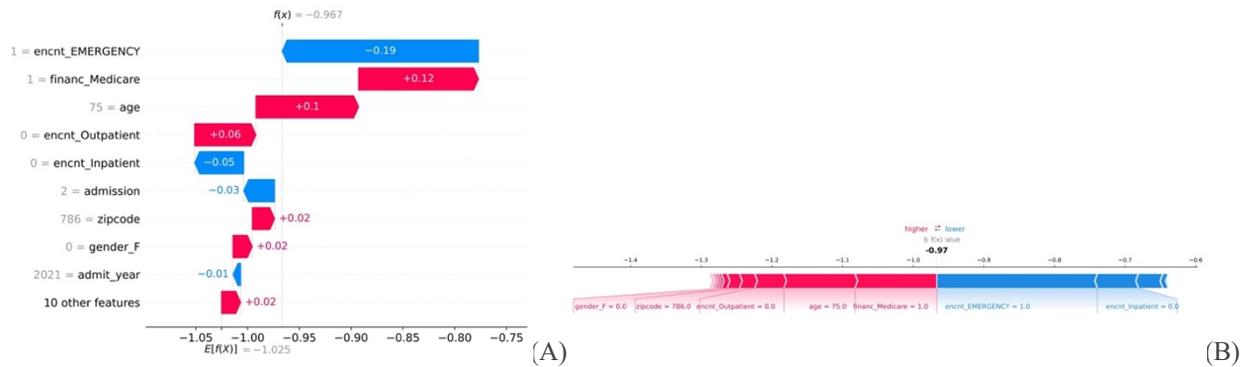

**Figure 2.** SHAP local interpretation (non-mortality).

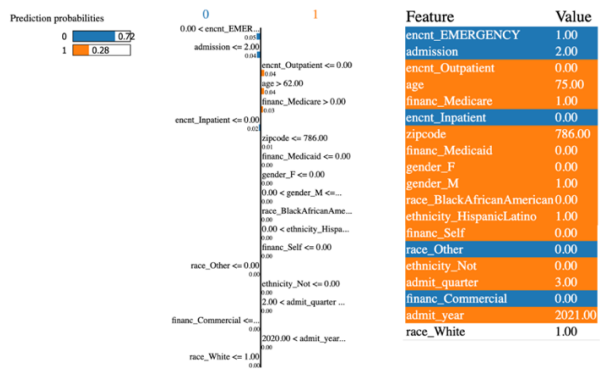

**Figure 3.** LIME local interpretation (non-mortality).

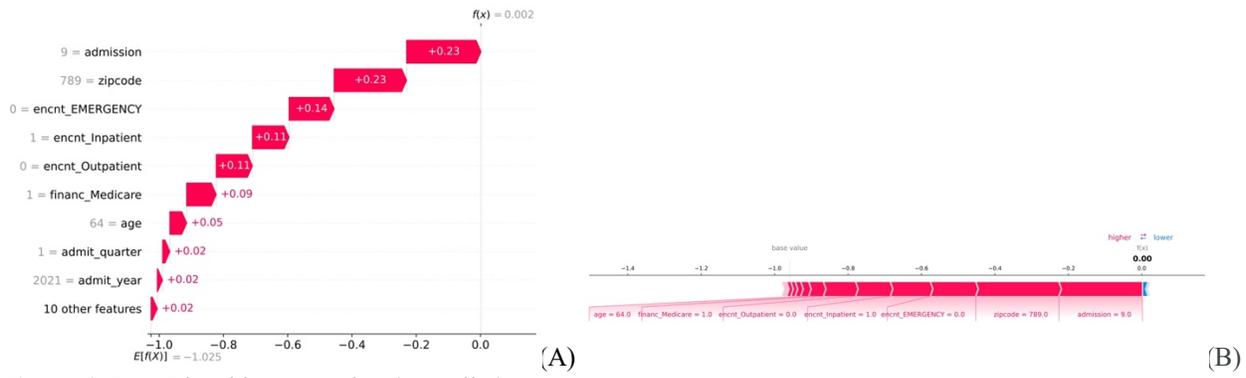

**Figure 4.** SHAP local interpretation (mortality).

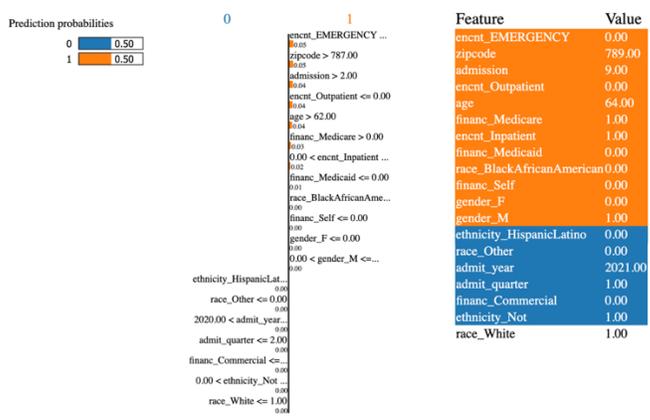

**Figure 5.** LIME local interpretation (mortality).

### 4.3 Cross validation

In the above two individual local interpretation, the contribution tendency toward non-mortality/ mortality and the rank of the impact of the features are quite consistent between SHAP and LIME. We therefore proceed to the cross validation of SHAP and LIME to assess the reliability of the local interpretations. Here we provide the descriptive statistics of impact value consistency and impact ranking difference in Table 3. As shown in the table, the mean ratio of impact value consistency is about 0.846 (SD=0.061), which means that in each individual local interpretation, on average about 85% of the features contribute the predictions to the same direction, either mortality or non-mortality. Among all the instances in the testing dataset, at least 63.2% and at most 94.7% of the feature impact values in SHAP and LIME are consistent. The mean of the impact ranking differences is around -0.032 (SD=2.207). Even though the difference of the ranking could be large, most (about 77.5%) of the impact ranking differences are between -2 to +2. Spearman's rank correlation was computed to assess the relationship between the SHAP ranking and LIME ranking of the features in local interpretation. For the features that are both contributing to mortality in SHAP and LIME, there was a strong, positive correlation between the two variables, which was statistically significant (r = 0.69, p=0.00). For the features that are both contributing to non-mortality in SHAP and LIME, there was also a strong, positive correlation between the two variables (r = 0.82, p=0.00).

**Table 3.** Descriptive statistics for feature consistency and ranking differences between SHAP and LIME.

|  | Mean | Median | Std | Minimum | Maximum |
|---|---|---|---|---|---|
| Ratio of Impact Value Consistency (per individual) | 0.846 | 0.842 | 0.061 | 0.632 | 0.947 |
| Impact Ranking Difference | -0.032 | 0 | 2.207 | -9 | 11 |

## 5 Discussion

In this research, we train an XGBoost model to predict the mortality due to COVID-19 with associated socio-economic factors. We employed the Shapley summary plot to demonstrate the decisive power of each feature. Furthermore, we performed a case-by-case patient prediction analysis using SHAP and LIME, which gives us insight into how the feature contribution differs between individuals. Finally, we cross-validated the local interpretation of SHAP and LIME using statistical methods in order to assess the reliability of the local interpretations.

In XAI interpretations, we could see several clear patterns, indicating that some features have a decisive impact on mortality prediction. When the encounter type is inpatient, the patient is at higher risk of mortality than the outpatient or emergency encounter type. This makes logical and possibly obvious sense considering only those with more severe symptoms would be admitted as an inpatient to the hospital vs. only being seen in the emergency department or having a short stay in the hospital where their status only meets the criteria for outpatient. This research also found that when the insurance class is Medicare, patients are at higher risk of mortality. While the insurance class is Medicaid, patients are more likely to survive. This is likely to happen because people with Medicare insurance are primarily older in age, who could have a higher likelihood of mortality. Previous research found that older adults have a higher mortality rate due to high Case Fatality Rate (CFR) and symptomatic infection rates [18,19]. Our findings showing clear patterns that older people are at higher mortality risk are consistent with the previous research. Additionally, the decisive power of gender also confirms previous research findings. Men are more at risk for worse outcomes and death because of the differences in sex hormones [20,21]. In contrast to [7], which mentioned that non-white race plays a vital role in southern states on the mortality prediction, our findings show that race has a shallow impact on the COVID-19 mortality compared to other factors. However, there are known limitations to representing the ground truth of race and ethnicity from EHR data [22]. Further work is needed to establish high-quality ground truth data reflecting social determinants of health.

SHAP global interpretations make it easier to understand the decisive power of the contribution of each feature in the model. The feature importance offers a holistic perspective of the impact of different feature values on the output. Observing the clustering patterns in the SHAP summary plot could help us interpret the clinical patterns on the relationship between socio-economic features and COVID-19 patients' mortality. In individual circumstances, differences in the feature differentially impact mortality between the individual and the average. Like other research using local interpretation, the factor of importance for individuals does not always align with the global interpretations. The top 5 impact features change as predicted for different cohorts [4]. In our research, for some patients, zip code and admission source could act as the top 2 impact feature, higher than encounter types and age, which is different from global interpretation. This inconsistency is valuable when we study individual circumstances and compare predictions of different classes. Many of the social determinants of health we studied in this research are non-modifiable. However, providing information to clinicians caring for these patients, particularly local interpretations, would provide insight as to why a prediction tool indicates increased risk of mortality for the given patient and could drive quality improvement initiatives or increased resource allocation when certain features are met, such as a patient who is transferred from another facility, is older, or has Medicaid. Applying these methods to additional features would support clinical and policy decisions with explainability of what features contribute the most to clinical predictions.

Even though SHAP has a more precise interpretation from the view of the computational mechanism, LIME local interpretation did not show significant differences in features consistency and rankings compared to SHAP in the COVID-19 patient mortality prediction. The cross-validation provides evidence for the reliability of XAI local interpretations. Furthermore, the local model assumptions method allows LIME to save substantial time for the local interpretation. Therefore, LIME is more efficient than SHAP in implementation. In some clinical circumstances requiring rapid reference content of the model prediction, LIME could be a potentially efficient and low-cost local interpretation tool to give preliminary results.

## 6 Conclusion

XAI explains the model prediction explicitly, including feature importance, decisive power, and local interpretations. A variety of XAI tools also offer the possibility of cross-validation. XAI tools have the potential to considerably enhance the transparency and trustworthiness of the prediction results. Health providers could utilize these Explainable AI tools to validate the AI prediction results and quickly make medical and health policy judgments and decisions. Possible future research includes utilizing other datasets to increase generalizability, conducting assessments of collinearity or correlation of features, as well as study the implement of these XAI tools into other clinical dataset to understand the application value in different clinical scenarios.

**Acknowledgement**
This research is supported by the COVID-19 Research Accelerator Grant funded by Gates Foundation (CORONAVIRUSHUB-D-21-00132). We thank Cole Maguire for his expertise and assistance throughout the data processing of our study and for the help in writing and reviewing the manuscript.